%% file: main.tex
\documentclass[10pt,twocolumn,letterpaper]{article}

 \usepackage{multirow}
 \usepackage{booktabs}
 \usepackage{amsmath}
 \usepackage{pifont}
 \usepackage{indentfirst} 
 \usepackage[pagenumbers]{cvpr} % To force page numbers, e.g. for an arXiv version

\input{preamble}

\definecolor{cvprblue}{rgb}{0.21,0.49,0.74}
\usepackage[pagebackref,breaklinks,colorlinks,citecolor=cvprblue]{hyperref}

\def\paperID{95} % *** Enter the Paper ID here
\def\confName{CVPR}
\def\confYear{2024}

\title{Multi-Level Feature Fusion Network for Lightweight Stereo Image Super-Resolution}

\author{Yunxiang Li, Wenbin Zou, Qiaomu Wei\\
Fuzhou University\\
Institution1 address\\
{\tt\small firstauthor@i1.org}
\and
Second Author\\
Institution2\\
First line of institution2 address\\
{\tt\small secondauthor@i2.org}
}

\author{Yunxiang Li$^{1, }\thanks{Equal contributions.}$, Wenbin Zou$^{2, *}$, Qiaomu Wei$^{3, *}$, Feng Huang$^{4,}\thanks{Corresponding author.}$, Jing Wu$^5$\\
Fuzhou University.$^{1,4,5}$ South China University of Technology.$^2$ \\Chengdu University of Information Technology.$^3$\\
{\tt\small 1033649629@qq.com, alexzou14@foxmail.com, 1642445844@qq.com,} \\
{\tt\small huangf@fzu.edu.cn, wujing@fzu.edu.cn}\\
}

\begin{document}
\maketitle
\input{sec/0_abstract}
\input{sec/1_intro}
\input{sec/2_related}

\input{sec/3_Multi_Level_Feature_Fusion_Network}

\input{sec/4_Experiments}
\input{sec/5_Conclusion}
{
    \small
    \bibliographystyle{ieeenat_fullname}
    \bibliography{main}
}


\end{document}

%% file: preamble.tex
\usepackage[dvipsnames]{xcolor}

%% file: sec/0_abstract.tex
\begin{abstract}
Stereo image super-resolution utilizes the cross-view complementary information brought by the disparity effect of left and right perspective images to reconstruct higher-quality images. Cascading feature extraction modules and cross-view feature interaction modules to make use of the information from stereo images is the focus of numerous methods. However, this adds a great deal of network parameters and structural redundancy. To facilitate the application of stereo image super-resolution in downstream tasks, we propose an efficient Multi-Level Feature Fusion Network for Lightweight Stereo Image Super-Resolution (MFFSSR). Specifically, MFFSSR utilizes the Hybrid Attention Feature Extraction Block (HAFEB) to extract multi-level intra-view features. Using the channel separation strategy, HAFEB can efficiently interact with the embedded cross-view interaction module. This structural configuration can efficiently mine features inside the view while improving the efficiency of cross-view information sharing. Hence, reconstruct image details and textures more accurately. Abundant experiments demonstrate the effectiveness of MFFSSR. We achieve superior performance with fewer parameters. The source code is available at \url{https://github.com/KarosLYX/MFFSSR}.
\end{abstract}

%% file: sec/1_intro.tex
\section{Introduction}
\label{sec:intro}

\begin{figure}
	\centering
	\includegraphics[width=8cm]{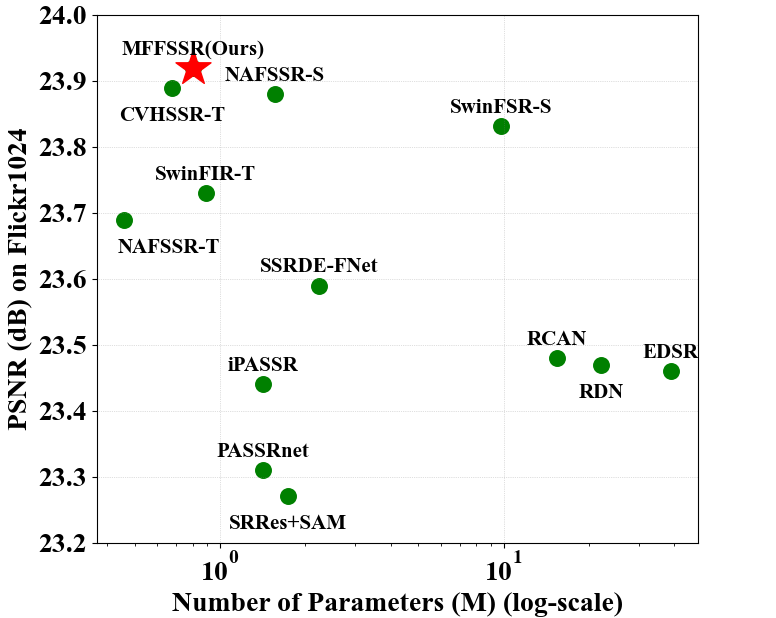}\\
        \caption{Comparison of the performance and complexity of state-of-the-art methods for $4 \times $ stereo SR on the Flickr1024 \cite{37} test set. Our MFFSSR achieves superior performance with fewer parameters.}
	% \caption{Parameters vs. PSNR of models for 2× stereo SR on Flickr1024 \cite{Flickr1024} test set. Our CVHSSR families achieve the state-of-the-art performance with fewer parameters.}
	\label{ComparsionMultiAdd}
\end{figure}

Stereo imaging utilizes two cameras to simulate the visual system of human, which has been widely applied in various fields such as augmented reality (AR) \cite{1}, virtual reality (VR) \cite{2}, and autonomous driving \cite{3}. However, the hardware costs of stereo imaging devices may lead to low resolution (LR). Image super-resolution (SR) can effectively enhance the perceived quality of images by restoring high-frequency details lost during the imaging process using computational optics methods, thus attracting widespread attention.

Recent years, the breakthrough of convolutional neural network (CNN) and Transformer technologies has enabled deep learning-based SR methods to demonstrate powerful performance in single image super-resolution (SISR) tasks \cite{4,5,6,7}. Different from SISR, which can only rebuild a high-resolution (HR) image with intra-view information, stereo images contain additional complementary information from cross-views. By fully leveraging the correlation between left and right perspective images, it is possible to reconstruct higher-quality HR images. However, the disparity effect can introduce uncertainty in the projected positions of objects across different perspective views. Positional variations are more pronounced for objects closer to the camera than for those farther away. This complexity makes it challenging to effectively harness details from stereo images. To address this issue, existing methods generally focus on designing complex networks and training strategies. For example, Jeon \textit{et al}. \cite{8} use parallax prior and a two-stage joint network to enhance the spatial resolution of stereo images. Wang \textit{et al}. \cite{9} introduced a parallax-attention mechanism and achieved feature fusion from cross-views based on similarity measurement. Recently, Chu \textit{et al}. \cite{10} refined the Nonlinear Activation Free Network, NAFNet \cite{11}, and adapted it to stereo scenarios. Cheng \textit{et al}. \cite{12} suggested a hybrid Transformer and CNN Attention Network among with a three-stage training strategy. They demonstrated excellent results in the NTIRE Stereo Image Super-Resolution Challenge held in 2022 and 2023 respectively.

Despite the capability achieved by the aforementioned methods in extracting information from stereo images, directly cascading feature extraction modules and cross-view feature interaction modules can lead to significant parameter and element redundancy, posing challenges for deployment on edge computing platforms. Therefore, exploring ways to reuse feature information for both intra-view and cross-view feature fusion is crucial for improving the efficiency of stereo image SR networks.

In this work, we develop a Multi-Level Feature Fusion Network for Lightweight Stereo Image Super-Resolution  (MFFSSR) to address the above issue. By using the channel separation strategy, we selectively learn intra-view and cross-view information, thereby reducing computational complexity. Specifically, we design a novel Hybrid Attention Feature Extraction Block (HAFEB) to extract multi-level intra-view features. We use Cross-View Interaction Module (CVIM) to extract cross-view information, which has been proven to be effective in \cite{13}. In addition, we embed CVIM within HAFEB and utilize a branching structure to enhance the efficiency of cross-view feature interaction. Through these structures, MFFSSR can effectively integrate multi-level features, achieving high-quality SR with fewer parameters.

The main contributions of this work are as follows:

\begin{itemize}[leftmargin=12pt, itemindent=12pt]
   
 \item We design HAFEB to extract and fuse multi-level intra-view features. By combining Channel Attention (CA) and Large Kernel Attention (LKA), HAFEB simultaneously reconstructs image details and structures while learning the correlations between local features. Residual connections further facilitate the transmission and fusion of features between different hierarchical levels, thereby preserving the richness and diversity of the extracted features.
   
 \item We integrate CVIM into HAFEB using a branching structure, leveraging partial intra-view features for cross-view interaction. Through a channel separation strategy, we optimize the cross-view information sharing mechanism, thus increasing efficiency and reducing computational complexity.
   
\item Based on the designed framework, we propose an effective and lightweight stereo image SR method. As shown in Figure \ref{ComparsionMultiAdd}, we achieve superior performance with fewer parameters. Extensive experiments confirm the effectiveness of our approach.
\end{itemize}

%--------------------------------------------------------------------------
\begin{figure*}[ht]
\vspace{-0.5cm}
\setlength{\abovecaptionskip}{0.1cm} 
\setlength{\belowcaptionskip}{-0.5cm}%调整caption与下文的距离
	\centering
	\includegraphics[width=0.8\textwidth]{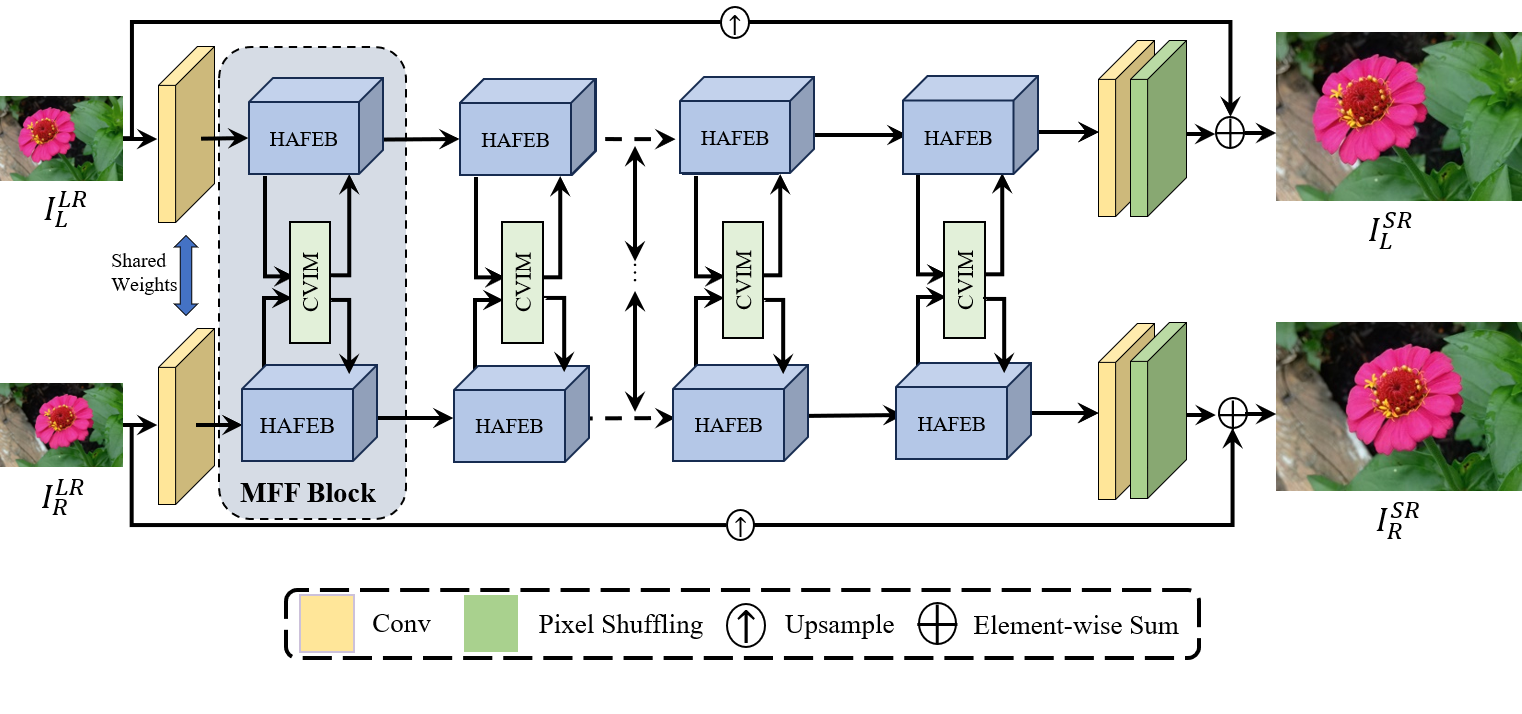}\\
	\caption{The framework of Multi-Level Feature Fusion Network for Lightweight Stereo Image SR (MFFSSR). HAFEB (shown in Figure \ref{MFFBlock}.) and CVIM (shown in Figure \ref{CVIM}.) represent the Hybrid Attention Feature Extraction Block and the Cross-View Interaction Module, respectively. Two HAFEBs with an embedded CVIM compose a MFF Block.}
	\label{frame}
\end{figure*}

%% file: sec/2_related.tex
{\section{Related Works}
\label{sec:related}}

% All text must be in a two-column format.
% The total allowable size of the text area is $6\frac78$ inches (17.46 cm) wide by $8\frac78$ inches (22.54 cm) high.
% Columns are to be $3\frac14$ inches (8.25 cm) wide, with a $\frac{5}{16}$ inch (0.8 cm) space between them.
% The main title (on the first page) should begin 1 inch (2.54 cm) from the top edge of the page.
% The second and following pages should begin 1 inch (2.54 cm) from the top edge.
% On all pages, the bottom margin should be $1\frac{1}{8}$ inches (2.86 cm) from the bottom edge of the page for $8.5 \times 11$-inch paper;
% for A4 paper, approximately $1\frac{5}{8}$ inches (4.13 cm) from the bottom edge of the
% page.

%------------------------------------------------------------------------
\subsection{Single Image Super-Resolution}

Recovering HR images from LR images is the aim of image SR, deep learning-based methods achieve this goal by learning the mapping relationship between a large number of LR images and their corresponding HR images. Since Dong \textit{et al}. \cite{14} initially suggested using CNN to achieve image SR task, many deep learning-based methods have emerged, demonstrating excellent performance. Kim \textit{et al}. \cite{15} further improved the effectiveness of CNN in image SR by increasing the depth of the network. By adding dense \cite{16,17} and residual \cite{18,19} connections, researchers have optimized the information flow and feature reuse amongst deep neural networks, thus increasing models’ robustness and training speed. However, images contain rich multi-level information, and different information contributes differentially to the image SR task. In order to focus more on the important features and structural information in the image,  Zhang \textit{et al}. \cite{19}  proposed the channel attention mechanism. Since then, various attention mechanisms \cite{20,21,22,23,24,25,26} have been proposed as effective means to enhance the performance of image SR. 

Recently, Transformer has achieved significant success in the field of computer vision, with Transformer-based SR models achieving state-of-the-art (SOTA) results. However, these models often have a large number of parameters, which limits their practical use. Additionally, single image super-resolution (SISR) can only utilize information within the image itself, without fully exploiting complementary information from other images, thereby restricting further enhancement in potency.

%Wang \textit{et al}.  \cite{9} introduced a Parallax Attention Module (PAM) with a global receptive field along the epipolar line to handle large disparity variation between different stereo images. Based on their work, Duan \textit{et al}. \cite{27} presented a Parallax-based Spatial and Channel Attention Module (PSCAM) to fully utilize the spatial and channel-wise properties to acquire the stereo correspondence. Yan \textit{et al}. \cite{28} proposed a method composed of three networks: a monocular network for spatial information extraction, a binocular network for cross-view information extraction, and a disparity flow network for view alignment.
\subsection{Stereo Image Super-Resolution}
Stereo image SR can make use of information across views to further enhance the resolution effect. Jeon \textit{et al}. \cite{8} proposed the first deep learning-based stereo image SR algorithm, StereoSR, which uses parallax prior and a two-stage joint network enhance the spatial resolution of stereo images. Song \textit{et al}. \cite{29} suggested a Self and Parallax Attention Mechanism (SPAM) to recover HR features while preserve stereo consistency between image pairs. In order to leverage texture-rich single image datasets, Ying \textit{et al}. \cite{30} developed a generic Stereo Attention Module (SAM) to extend SISR network to a stereo image SR network. Xu \textit{et al}. \cite{31} incorporated the idea of bilateral grid processing into a CNN framework to effectively utilize cross-view information. Wang \textit{et al}. \cite{32} utilized a Bi-directional Parallax Attention Module (BiPAM) to simultaneously interact with information from both left and right perspective images. Additionally, they addressed the issue of inconsistent illumination between left and right perspective images in real-world scenes by improving the loss function. Chu \textit{et al}. \cite{10} used a stack of NAFBlock \cite{11} for intra-view feature extraction and combined it with stereo cross-attention modules for cross-view feature interaction, resulting in excellent performance. Zou \textit{et al}. \cite{13} improved upon their work by designing the CVHSSR, which effectively conveys mutual information between different views. In addition, Transformer-based methods \cite{12,33,34} have begun to be applied in the field of stereo image SR, achieving impressive outcomes.

However, the above methods often focus solely on performance while neglecting the potential for application in downstream tasks. To address this issue, we design a lightweight stereo image SR network, redefining the processes of intra-view feature extraction and cross-view feature interaction to enhance the efficiency of the network.

%% file: sec/3_Multi_Level_Feature_Fusion_Network.tex
\section{Multi-Level Feature Fusion Network}
\label{sec:3}

\subsection{Overall Framework}

\begin{figure*}[ht]
\vspace{-0.3cm}
\setlength{\abovecaptionskip}{0.1cm} 
\setlength{\belowcaptionskip}{-0.5cm}%调整caption与下文的距离
	\centering
	\includegraphics[width=0.8\textwidth]{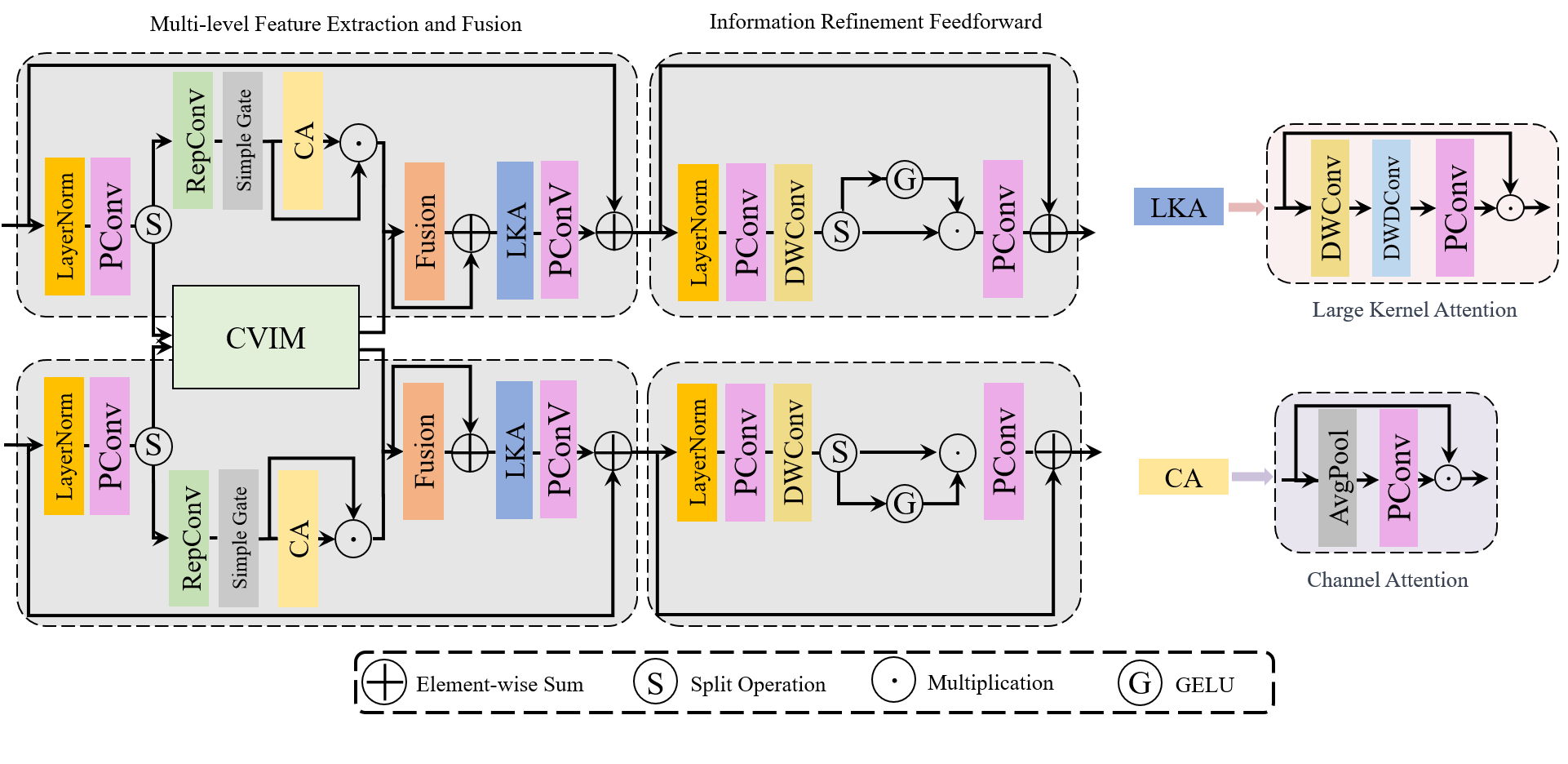}\\
	\caption{The architecture of our proposed Multi-level Feature Fusion Block (MFF Block). A MFF Block consists of two Hybrid Attention Feature Extraction Blocks (HAFEB) and an embedded Cross-View Interaction Module (CVIM). Each HAFEB has two components: Multi-level Feature Extraction and Fusion (MFEF) and Information Refinement Feedforward (IRF). The HAFEBs for the left and right views are connected to CVIM through the branch structures in MFEF, facilitating the interaction and fusion of features across the views. PConv, RepConv, DWConv and DWDConv in the figure represent point-wise convolution, reparameterized convolution, depth-wise convolution, and depth-wise dilation convolution, respectively.
}
	\label{MFFBlock}
\end{figure*}

The network proposed by us is illustrated in Figure \ref{frame}. MFFSSR employs a dual-branch network with shared weights to restore images from both left and right perspectives. It consists of three parts: shallow feature extraction, deep feature extraction and interaction, and stereo image reconstruction. The Multi-Level Feature Fusion Block (MFF Block) is the core component of the deep feature extraction and interaction, which consists of two Hybrid Attention Feature Extraction Blocks (HAFEBs) and an embedded Cross-View Interaction Module (CVIM). Detailed information about HAFEB and CVIM will be presented in Section \ref{sec:3.2} and Section \ref{sec:3.3}, respectively. Specifically, the operation process of MFFSSR is as follows.

Firstly, given a pair of LR stereo images $I_L^{LR},I_R^{LR} \in {R^{H \times W \times 3}}$, a simple convolutional operation is used for them to extract the shallow features $F_L^S,F_R^S \in {R^{H \times W \times C}}$, where \textit{H}, \textit{W}, and \textit{C} represent the image’s height, width, and number of channels, respectively. This process can be described as:
\begin{equation}
F_{L,R}^S = {H_{{\rm{conv}}}}(I_L^{LR},I_R^{LR})
\end{equation}
where ${H_{{\rm{conv}}}}$ denotes $3 \times 3$ convolution operation.

Next, we perform deep feature extraction and interactive fusion using MFF Block on the acquired features. The number of MFF Blocks, denoted by \textit{N}, is flexible and can be adjusted. This process can be described as:
\begin{equation}
F_{L,R}^D = H_{{\rm{MFF}}}^N(H_{{\rm{MFF}}}^{N - 1}( \cdot  \cdot  \cdot (H_{{\rm{MFF}}}^1(F_{L,R}^S))))
\end{equation}
\begin{equation}
F_{L,R}^{i + 1} = H_{{\rm{MFF}}}^{}(F_{L,R}^i)
\end{equation}
where ${H_{{\rm{MFF}}}}$ denotes MFF Block. $F_{L,R}^D, F_{L,R}^{i + 1}$ 
denote the features after deep extraction and interactive fusion and the features obtained after processing by the $i$-th MFF Block, respectively.

Finally, we utilize the pixel shuffling operation to upsample the output features to the HR size. Furthermore, a global residual structure is used to maintain input image features and increase the performance of SR. This process can be described as:
\begin{equation}
I_L^{SR} = H_{up}^{}(F_L^D) + H_{up}^{}(I_L^{LR})
\end{equation}
\begin{equation}
I_R^{SR} = H_{up}^{}(F_R^D) + H_{up}^{}(I_R^{LR})
\end{equation}
where $H_{up}^{}$ denotes upsampling operation. $I_L^{SR}$ and $I_R^{SR}$ represent the final left and right perspective images after SR, respectively.

\subsection{Intra-View Feature Extraction}
\label{sec:3.2}
Stereo images contain information spanning global, local, and cross-view ranges. Intra-view feature extraction serves as the foundation for cross-view interaction. To efficiently capture and fuse these multi-level features, we introduce the  Hybrid Attention Feature Extraction Block (HAFEB).

As shown in Figure \ref{MFFBlock}, the HAFEB consists of two components: (1) Multi-level Feature Extraction and Fusion (MFEF) and (2) Information Refinement Feedforward (IRF). In addition to channel attention and large kernel attention mechanisms, we also employ reparameterized convolution (RepConv) in MFEF. Their comprehensive use significantly enhances the capability and flexibility of HAFEB in feature extraction. Furthermore, we use branch structures to interact partial intra-view features with the embedded Cross-View Interaction Module (CVIM), therefore reducing computational complexity and substantially improving the efficiency of cross-view information fusion.

Given an input tensor $F{\rm{in}} \in {R^{H \times W \times C}}$, the working process of MFEF can be described as follows:

\begin{equation}
{\begin{aligned}
{F_{{\rm{MFEF}}}} &= H_{{\rm{pconv}}}^2({H_{{\rm{LKA}}}}(H_{\rm{f}}^{}(\kappa 
(H_{{\rm{pconv}}}^1(LN({F_{{\rm{in}}}})))) \\
&+ \kappa (H_{{\rm{pconv}}}^1(LN({F_{{\rm{in}}}}))))) + {F_{{\rm{in}}}}
\end{aligned}}
\end{equation}
where $LN( \cdot )$ denotes layer normalization. $H_{{\rm{pconv}}}^{( \cdot )}$, $H_{{\rm{f}}}$, $H_{{\rm{LKA}}}$ represent $1 \times 1$ point-wise convolution, feature fusion operation, and large kernel attention, respectively. ${F_{{\rm{MFEF}}}}$ is the output feature of MFEF. We use notation $\kappa ( \cdot )$ to represent hybrid feature fusion extraction operation. Specifically, given the input feature $X \in {R^{H \times W \times C}}$ , it is firstly split into two parts ${X_1} \in {R^{H \times W \times \theta }}$, $X_2 \in {R^{H \times W \times (1 - \theta )}}(\theta  \in [0,1])$ on channel dimension. $\lambda $ is a hyperparameter that controls the ratio. We set $\theta  = 0.75$  to balance between the parameters and efficiency. More detailed information about it can be found in the ablation study in Section \ref{sec:4.3}. Then, $X_1$ and $X_2$ are used for further feature extraction and cross-view information interaction, respectively. This process can be described as:
\begin{equation}
{\begin{aligned}
\kappa (X) &= {\delta _{{\rm{SG}}}}({H_{{\rm{rconv}}}}(X_1)) \odot {H_{{\rm{CA}}}}({\delta _{{\rm{SG}}}}({H_{{\rm{rconv}}}}(X_1)))\\
&+ {F_{{\rm{CVIM}}}}(X_2)
\end{aligned}}
\end{equation}
where ${H_{{\rm{rconv}}}}$ and ${F_{{\rm{CVIM}}}}$   represent the RepConv and the output feature of CVIM, respectively. ${\delta _{{\rm{SG}}}}$ means SimpleGate function and $\odot$ denotes element-wise multiplication.

The internal computation process of large kernel attention and channel attention can be described as follows:
\begin{equation}
{H_{{\rm{LKA}}}}(X) = X \odot ({H_{{\rm{pconv}}}}({H_{{\rm{dd7}}}}({H_{{\rm{d5}}}}(X))))
\end{equation}
\begin{equation}
{H_{{\rm{CA}}}}(X) = X \odot ({H_{{\rm{pconv}}}}({H_{{\rm{Avg}}}}(X)))
\end{equation}
where ${H_{{\rm{Avg}}}}$, ${H_{{\rm{d5}}}}$ and ${H_{{\rm{dd7}}}}$  represent average pooling operation, $5 \times 5$ depth-wise convolution, and $7 \times 7$ depth-wise dilation convolution respectively.

Then, IRF employs a non-linear gate mechanism to focus on complementary details across different levels. The working process of IRF can be described as follows:
\begin{equation}
\begin{aligned}
{F_{{\rm{out}}}} &= H_{{\rm{pconv}}}^4({\delta _{{\rm{NG}}}}(H_{{\rm{d3}}}^1(H_{{\rm{pconv}}}^3(LN({F_{{\rm{MFEF}}}})))))\\ &+ {F_{{\rm{MFEF}}}}
\end{aligned}
\end{equation}
where $H_{{\rm{d3}}}^{\left(  \cdot  \right)}$ and ${\delta _{{\rm{NG}}}}$ represent $3 \times 3$ depth-wise convolution and non-linear gate function, respectively. The ${F_{{\rm{out}}}}$ denotes the output feature of HAFEB.

% \begin{figure}[t]
%  \centering
%  \includegraphics[width=\columnwidth]{CHIMB.pdf}
%  \caption{The architecture of our proposed cross-hierarchy information mining block (CHIMB). PConv, DWConv and DWDConv in the figure represent point-wise convolution, depth-wise convolution, and depth-wise dilation convolution, respectively.}\label{fig: CHIMB}
%  \label{CHIMB}
%  \end{figure}

\begin{figure}[t]
 \centering
 \includegraphics[width=\columnwidth]{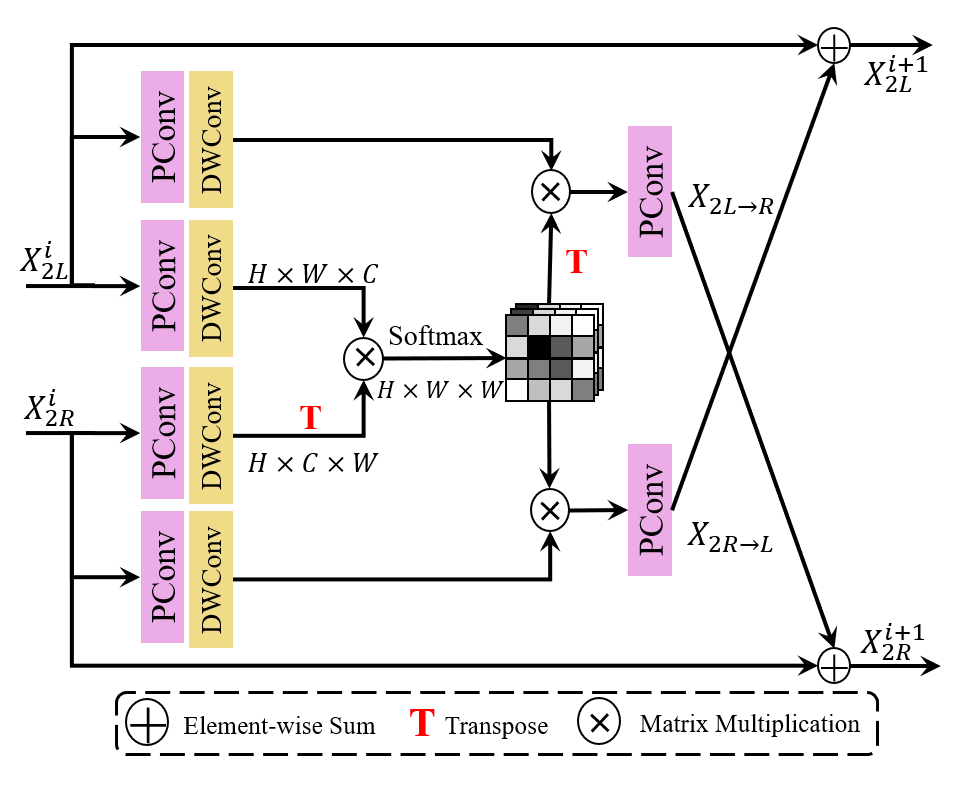}
 \caption{The architecture of Cross-View Interaction Module (CVIM). It is embedded in two Hybrid Attention
Feature Extraction Blocks of the parallel branches to achieve efficient cross-view feature interaction. PConv, DWConv in the figure represent point-wise convolution, depth-wise convolution,  respectively.}
 \label{CVIM}
 \end{figure}

\begin{table*}
\caption{Quantitative results achieved by different methods on the KITTI2012 \cite{10}, KITTI2015 \cite{24}, Middlebury \cite{27}, and Flickr1024 \cite{33} test sets. Params represents the number of parameters of the networks. Here, PSNR/SSIM values achieved on both the left images (i.e., \textit{Left}) and a pair of stereo images (i.e., $(Left + Right)/2$) are reported. The best and second best results are \textcolor{red}{red} and \textcolor{blue}{blue}.}
\label{tab:my-table}
\centering
\resizebox{16cm}{!}{%
\begin{tabular}{llc|ccc|cccc}
\bottomrule[1.2pt]
\multirow{2}{*}{Method} & \multicolumn{1}{c}{\multirow{2}{*}{Scale}} & \multirow{2}{*}{Params} & \multicolumn{3}{c|}{$Left$}                  & \multicolumn{4}{c}{$(Left+Right)/2$}                        \\ \cline{4-10} 
                        & \multicolumn{1}{c}{}                       &                         & KITTI2012    & KITTI2015    & Middlebury   & KITTI2012    & KITTI2015    & Middlebury   & Flickr1024   \\ \hline
VDSR \cite{15}        & $\times 2$                                 & 0.66M                   & 30.17/0.9062 & 28.99/0.9038 & 32.66/0.9101 & 30.30/0.9089 & 29.78/0.9150 & 32.77/0.9102 & 25.60/0.8534 \\
EDSR \cite{18}        & $\times 2$                                 & 38.6M                   & 30.83/0.9199 & 29.94/0.9231 & 34.84/0.9489 & 30.96/0.9228 & 30.73/0.9335 & 34.95/0.9492 & 28.66/0.9087 \\
RDN  \cite{7}         & $\times 2$                                 & 22.0M                   & 30.81/0.9197 & 29.91/0.9224 & 34.85/0.9488 & 30.94/0.9227 & 30.70/0.9330 & 34.94/0.9491 & 28.64/0.9084 \\
RCAN \cite{19}        & $\times 2$                                 & 15.3M                   & 30.88/0.9202 & 29.97/0.9231 & 34.80/0.9482 & 31.02/0.9232 & 30.77/0.9336 & 34.90/0.9486 & 28.63/0.9082 \\
StereoSR \cite{8}& $\times 2$                                 & 1.08M                   & 29.42/0.9040 & 28.53/0.9038 & 33.15/0.9343 & 29.51/0.9073 & 29.33/0.9168 & 33.23/0.9348 & 25.96/0.8599 \\
PASSRnet \cite{9}& $\times 2$                                 & 1.37M                   & 30.68/0.9159 & 29.81/0.9191 & 34.13/0.9421 & 30.81/0.9190 & 30.60/0.9300 & 34.23/0.9422 & 28.38/0.9038 \\
IMSSRnet \cite{42}& $\times 2$                                 & 6.84M                   & 30.90/-      & 29.97/-      & 34.66/-      & 30.92/-      & 30.66/-      & 34.67/-      & -/-          \\
iPASSR  \cite{32}   & $\times 2$                                 & 1.37M                   & 30.97/0.9210 & 30.01/0.9234 & 34.41/0.9454 & 31.11/0.9240 & 30.81/0.9340 & 34.51/0.9454 & 28.60/0.9097 \\
SSRDE-FNet \cite{36} & $\times 2$                                 & 2.10M                   & 31.08/0.9224 & 30.10/0.9245 & 35.02/0.9508 & 31.23/0.9254 & 30.90/0.9352 & 35.09/0.9511 & 28.85/0.9132 \\
PFT-SSR \cite{43}      & $\times 2$                                 & -                       & 31.15/0.9166 & 30.16/0.9187 & 35.08/0.9516 & 31.29/0.9195 & 30.96/0.9306 & 35.21/0.9520 & 29.05/0.9049 \\
SwinFIR-T \cite{5}& $\times 2$                                 & 0.89M                   & 31.09/0.9226 & 30.17/0.9258 & 35.00/0.9491 & 31.22/0.9254 & 30.96/0.9359 & 35.11/0.9497 & 29.03/0.9134 \\
NAFSSR-T \cite{10} & $\times 2$                                  & 0.45M                   & 31.12/0.9224 & 30.19/0.9253 & 34.93/0.9495 & 31.26/0.9254 & 30.99/0.9355 & 35.01/0.9495 & 28.94/0.9128 \\
NAFSSR-S \cite{10} & $\times 2$                                  & 1.54M                   & 31.23/0.9236 & 30.28/0.9266 & 35.23/0.9515 & 31.38/0.9266 & 31.08/0.9367 & 35.30/0.9514 & 29.19/0.9160 \\
CVHSSR-T \cite{13} & $\times 2$                                  & 0.66M                   & \textcolor{blue}{31.31}/\textcolor{blue}{0.9250} & \textcolor{blue}{30.33}/\textcolor{blue}{0.9277} & \textcolor{blue}{35.41}/\textcolor{red}{0.9533} & \textcolor{blue}{31.46}/\textcolor{blue}{0.9280} & \textcolor{blue}{31.13}/\textcolor{blue}{0.9377} & \textcolor{blue}{35.47}/\textcolor{red}{0.9532} & \textcolor{blue}{29.26}/\textcolor{blue}{0.9180} \\ 
MFFSSR (Ours)        & $\times 2$                                 &0.78M                    & \textcolor{red}{31.35}/\textcolor{red}{0.9255} & \textcolor{red}{30.36}/\textcolor{red}{0.9281} &         \textcolor{red}{35.45}/\textcolor{red}{0.9533} & \textcolor{red}{31.50}/\textcolor{red}{0.9285} & \textcolor{red}{31.16}/\textcolor{red}{0.9380} &\textcolor{red}{35.51}/\textcolor{blue}{0.9531} & \textcolor{red}{29.38}/\textcolor{red}{0.9198}  \\ \hline \hline
VDSR \cite{15}       & $\times 4$                                 & 0.66M                   & 25.54/0.7662 & 24.68/0.7456 & 27.60/0.7933 & 25.60/0.7722 & 25.32/0.7703 & 27.69/0.7941 & 22.46/0.6718 \\
EDSR \cite{18}       & $\times 4$                                 & 38.9M                   & 26.26/0.7954 & 25.38/0.7811 & 29.15/0.8383 & 26.35/0.8015 & 26.04/0.8039 & 29.23/0.8397 & 23.46/0.7285 \\
RDN \cite{7}         & $\times 4$                                 & 22.0M                   & 26.23/0.7952 & 25.37/0.7813 & 29.15/0.8387 & 26.32/0.8014 & 26.04/0.8043 & 29.27/0.8404 & 23.47/0.7295 \\
RCAN \cite{19}       & $\times 4$                                 & 15.4M                   & 26.36/0.7968 & 25.53/0.7836 & 29.20/0.8381 & 26.44/0.8029 & 26.22/0.8068 & 29.30/0.8397 & 23.48/0.7286 \\
StereoSR \cite{8}   & $\times 4$                                 & 1.42M                   & 24.49/0.7502 & 23.67/0.7273 & 27.70/0.8036 & 24.53/0.7555 & 24.21/0.7511 & 27.64/0.8022 & 21.70/0.6460 \\
PASSRnet \cite{9}& $\times 4$                                 & 1.42M                   & 26.26/0.7919 & 25.41/0.7772 & 28.61/0.8232 & 26.34/0.7981 & 26.08/0.8002 & 28.72/0.8236 & 23.31/0.7195 \\
SRRes+SAM \cite{30} & $\times 4$                                 & 1.73M                   & 26.35/0.7957 & 25.55/0.7825 & 28.76/0.8287 & 26.44/0.8018 & 26.22/0.8054 & 28.83/0.8290 & 23.27/0.7233 \\
IMSSRnet \cite{42}& $\times 4$                                 & 6.89M                   & 26.44/-      & 25.59/-      & 29.02/-      & 26.43/-      & 26.20/-      & 29.02/-      & -/-          \\
iPASSR \cite{32}    & $\times 4$                                 & 1.42M                   & 26.47/0.7993 & 25.61/0.7850 & 29.07/0.8363 & 26.56/0.8053 & 26.32/0.8084 & 29.16/0.8367 & 23.44/0.7287 \\
SSRDE-FNet \cite{36} & $\times 4$                                 & 2.24M                   & 26.61/0.8028 & 25.74/0.7884 & 29.29/0.8407 & 26.70/0.8082 & 26.43/0.8118 & 29.38/0.8411 & 23.59/0.7352 \\
PFT-SSR \cite{43}      & $\times 4$                                 & -                       & 26.64/0.7913 & 25.76/0.7775 & 29.58/0.8418 & 26.77/0.7998 & 26.54/0.8083 & 29.74/0.8426 & 23.89/0.7277 \\
SwinFIR-T \cite{5}& $\times 4$                                 & 0.89M                   & 26.59/0.8017 & 25.78/0.7904 & 29.36/0.8409 & 26.68/0.8081 & 26.51/0.8135 & 29.48/0.8426 & 23.73/0.7400 \\
NAFSSR-T \cite{10}  & $\times 4$                                 & 0.46M                   & 26.69/0.8045 & 25.90/0.7930 & 29.22/0.8403 & 26.79/0.8105 & 26.62/0.8159 & 29.32/0.8409 & 23.69/0.7384 \\
NAFSSR-S \cite{10}  & $\times 4$                                 & 1.56M                   & 26.84/0.8086 & 26.03/0.7978 & 29.62/0.8482 & 26.93/0.8145 & 26.76/0.8203 & 29.72/0.8490 & 23.88/0.7468 \\ 
CVHSSR-T \cite{13}  & $\times 4$                                 & 0.68M                   & \textcolor{blue}{26.88}/\textcolor{blue}{0.8105} & \textcolor{red}{26.03}/\textcolor{blue}{0.7991} & \textcolor{blue}{29.62}/\textcolor{blue}{0.8496} & \textcolor{blue}{26.98}/\textcolor{blue}{0.8165} & \textcolor{red}{26.78}/\textcolor{blue}{0.8218} & \textcolor{blue}{29.74}/\textcolor{blue}{0.8505} & \textcolor{blue}{23.89}/\textcolor{blue}{0.7484} \\ 
MFFSSR (Ours)         & $\times 4$                                 & 0.84M                   & \textcolor{red}{26.89}/\textcolor{red}{0.8109} & \textcolor{red}{26.05}/\textcolor{red}{0.7992} &        \textcolor{red}{29.64}/\textcolor{red}{0.8498} & \textcolor{red}{26.99}/\textcolor{red}{0.8169} & \textcolor{red}{26.78}/\textcolor{red}{0.8219} &\textcolor{red}{29.75}/\textcolor{red}{0.8507} & \textcolor{red}{23.92}/\textcolor{red}{0.7503} \\ \toprule[1.2pt]
\end{tabular}%
}
\label{Quantitative Compare}
\end{table*}
 
\subsection{Cross-View Feature Interaction}
\label{sec:3.3}
We refine the  Cross-View Interaction Module (CVIM) proposed in CVHSSR \cite{13}. Redundant cross-view feature interaction has little contribution to SR performance improvement but can lead to a significant increase in computational complexity. To improve cross-view interaction efficiency and reduce parameters, we constrain the input feature in dimension. Additionally, layer normalization is removed for better integration into the Hybrid Attention Feature Extraction Blocks. The details of CVIM is as shown in Figure \ref{CVIM}. It combines Scaled DotProduct Attention \cite{35}, which utilizes queries and keys to generate corresponding weights: 
\begin{equation}
Attention({\bf{Q}},{\bf{K}},{\bf{V}}) = softmax({\bf{Q}}{{\bf{K}}^T}/\sqrt C ){\bf{V}}
\end{equation}
where ${\bf{Q}} \in {R^{H \times W \times C}}$ is the query matrix from one view, and ${\bf{K}},{\bf{V}} \in {R^{H \times W \times C}}$  are key and query matrices to another view.

CVIM efficiently facilitates the interaction between left and right view information. Given the input stereo partial intra-view features $X_{2L}^i,X_{2R}^i \in {R^{H \times W \times C}}$, we can get the cross-view fusion features $X_{2L \to R}^{}$ through the following process:
\begin{equation}
{{\bf{Q}}_{\rm{L}}} = H_{d3}^{{Q_L}}(H_{{\rm{pconv}}}^{{Q_L}}(X_{2L}^i))
\end{equation}
\begin{equation}
{{\bf{K}}_{\rm{R}}} = H_{d3}^{{K_R}}(H_{{\rm{pconv}}}^{K_{R_{}}}(X_{2R}^i))
\end{equation}
\begin{equation}
{{\bf{V}}_{\rm{R}}} = H_{d3}^{{V_R}}(H_{{\rm{pconv}}}^{V_{R_{}}}(X_{2R}^i))
\end{equation}
\begin{equation}
X_{2L \to R}^{} = H_{{\rm{pconv}}}^RAttentio{n_{{\rm{L}} \to {\rm{R}}}}({{\bf{Q}}_{\rm{L}}},{{\bf{K}}_{\rm{R}}},{{\bf{V}}_{\rm{R}}})
\end{equation}

$X_{2R \to L}^{}$ can be obtained through the similar process:
\begin{equation}
{{\bf{Q}}_{\rm{R}}} = H_{d3}^{{Q_R}}(H_{{\rm{pconv}}}^{{Q_R}}(X_{2R}^i))
\end{equation}
\begin{equation}
{{\bf{K}}_{\rm{L}}} = H_{d3}^{{K_L}}(H_{{\rm{pconv}}}^{{K_L}}(X_{2L}^i))
\end{equation}
\begin{equation}
{{\bf{V}}_{\rm{L}}} = H_{d3}^{{V_L}}(H_{{\rm{pconv}}}^{{V_L}}(X_{2L}^i))
\end{equation}
\begin{equation}
X_{2R \to L}^{} = H_{{\rm{pconv}}}^LAttentio{n_{{\rm{R}} \to {\rm{L}}}}({{\bf{Q}}_{\rm{R}}},{{\bf{K}}_{\rm{L}}},{{\bf{V}}_{\rm{L}}})
\end{equation}

The cross and intra view features are finally fused to generate the output features $X_{2L}^{i + 1}$ and $X_{2R}^{i + 1}$:
\begin{equation}
X_{2L}^{i + 1} = {\gamma _L}X_{2L \to R}^{} + X_{2L}^i
\end{equation}
\begin{equation}
X_{2R}^{i + 1} = {\gamma _R}X_{2R \to L}^{} + X_{2R}^i
\end{equation}
where ${\gamma _L}$ and ${\gamma _R}$ are trainable channel-wise scales and initialized with zeros for stabilizing training.

% \begin{table}
% \caption{Ablation experiments of different components. We use PSNR and SSIM to perform quantitative comparison.}
% \label{tab:2}
% \centering
% \resizebox{7cm}{!}{%
% \begin{tabular}{c|cccc}
% \bottomrule[1.2pt]
% Method       &PSNR       &Params        &FLOPs      \\\hline
% SSRDE-FNet   &23.59      &2.24M         &427.265G   \\
% NAFSSR-S     &23.88      &1.54M         &36.531G     \\
% SwinFIRSSR   & -           &0.89M         &48.956G      \\
% SCGLANet     &-&0.75M      &28.244G       \\\hline
% MFFSSR(Ours)  &23.92     &0.91M      &27.415G  \\ \toprule[1.2pt]
% \end{tabular}%
% }
% \end{table}
\begin{table}
\caption{Efficiency evaluations with the state-of-the-art methods. Params represents the number of parameters of the network. \textit{v} represents the variant after reducing parameters. We use Params and FLOPs to evaluate efficiency.}
\label{tab:2}
\centering
\resizebox{5cm}{!}{%
\begin{tabular}{c|cccc}
\bottomrule[1.2pt]
Method             &Params        &FLOPs      \\\hline
NAFSSR-S \cite{10}            &1.54M         &36.531G     \\
SwinFIRSSR-\textit{v }            &0.89M         &48.956G      \\
SCGLANet-\textit{v}     &0.75M      &28.244G       \\\hline
MFFSSR (Ours)      &0.91M      &27.415G  \\ \toprule[1.2pt]
\end{tabular}%
}
\end{table}

\subsection{Loss Function}

The loss function defines the optimization objective of the SR network and plays a crucial role in determining how well it performs. Zou \textit{et al}. have already demonstrated the effectiveness of utilizing spatial and frequency domain losses to jointly guide the SR network for image restoration \cite{13}. 

Specifically, we use the MSE loss to measure the spatial structural difference between the SR images $I_{L,R}^{SR}$ and the HR images $I_{L,R}^{HR}$, which can be described as:
\begin{equation}
{L_{MSE}} = {1 \over N}{\sum\limits_{i = 1}^N {\left\| {I_{L,R}^{HR} - I_{L,R}^{SR}} \right\|} ^2}
\end{equation}

Additionally, frequency Charbonnier loss is introduced to guide the learning of high-frequency information in SR images, aiding in better preservation of details and textures. It can be defined as:
\begin{equation}
{L_{FC}} = {1 \over N}\sum\limits_{i = 1}^N {\sqrt {{{\left\| {FFT(I_{L,R}^{HR}) - FFT(I_{L,R}^{SR})} \right\|}^2} + {\varepsilon ^2}} }
\end{equation}
where $\varepsilon$ is a constant and is set to ${10^{ - 3}}$. $FFT\left(  \cdot  \right)$ denotes the fast Fourier transform.

In conclusion, the overall loss function can be expressed as:
\begin{equation}
{L_{Total}} = {L_{MSE}}(I_{L,R}^{HR},I_{L,R}^{SR}) + \lambda {L_{FC}}(I_{L,R}^{HR},I_{L,R}^{SR})
\end{equation}
where $\lambda $ is a hyperparameter, it is set to 0.01 to control the proportion of the frequency Charbonnier loss function.

%% file: sec/4_Experiments.tex
\section{Experiments}
\label{sec:4}

\begin{figure*}
	\centering %15cm
	\resizebox{14cm}{!}{\includegraphics[]{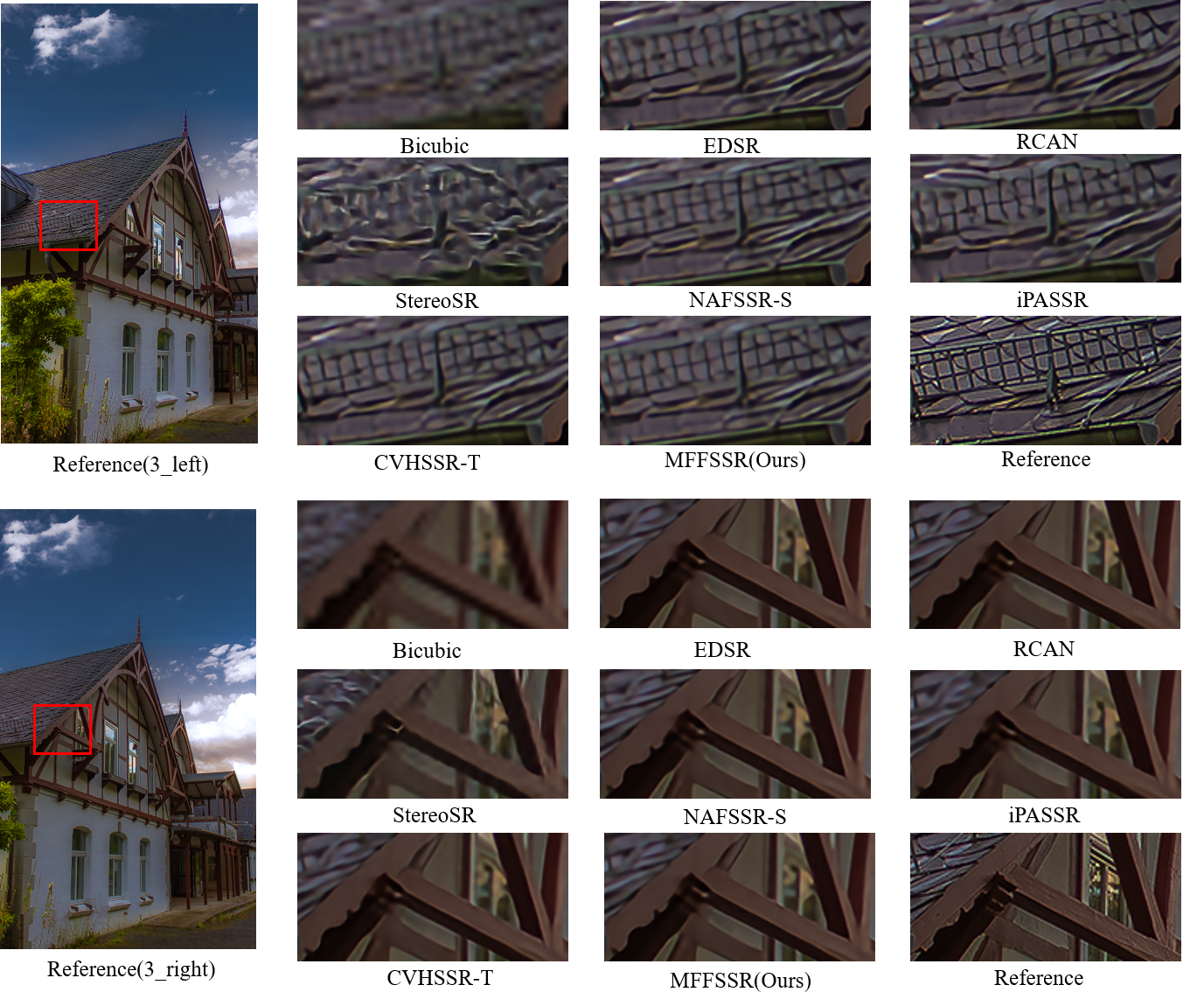}}\\
	\caption{Visual results ($\times 4 $ SR) achieved by different methods on the Flickr1024 \cite{37} test set.}
	\label{cpimageflickr}
\end{figure*}

\begin{figure*}
	\centering %15cm
	\resizebox{14cm}{!}{\includegraphics[]{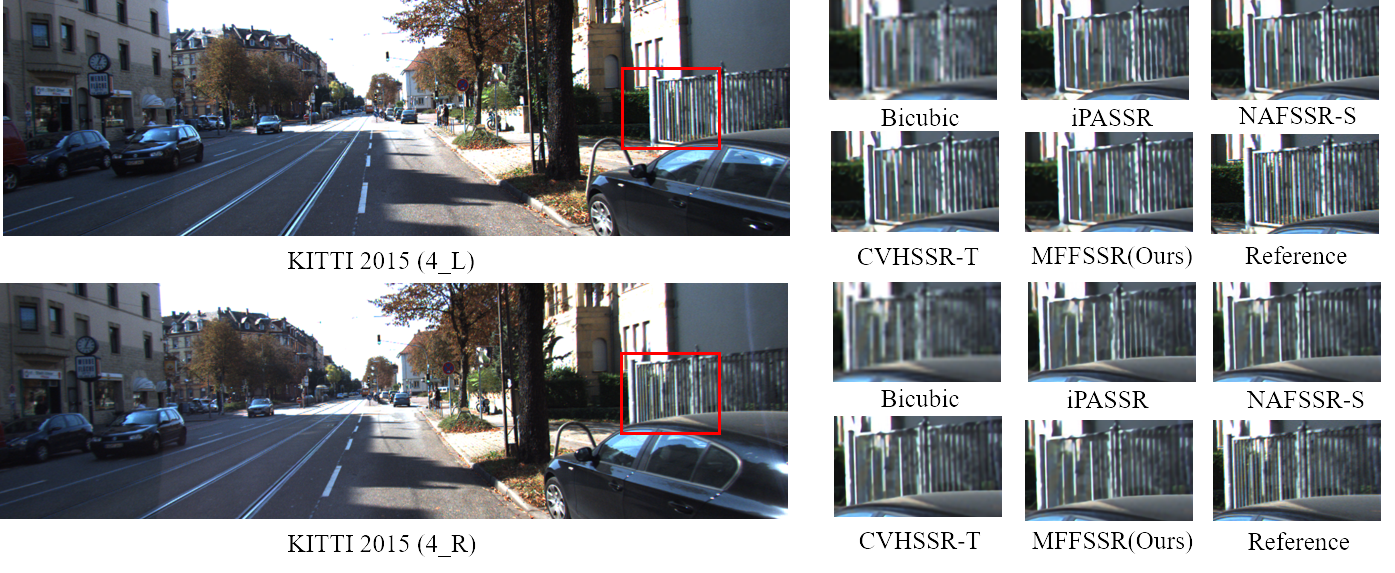}}\\
	\caption{Visual results ($\times 4 $ SR) achieved by different methods on the KITTI 2015 \cite{40}  test set.}
	\label{cpimagekitti}
\end{figure*}

\subsection{Implementation Details}
In this section, we provide a detailed overview of the datasets, evaluation metrics, and model configurations.

\textbf{Datasets.} Following previous works \cite{30,31,32,36}, we used publicly available stereo image datasets for our training and testing. Specifically, we used 800 pairs of images from the Flickr1024 \cite{37} dataset and 60 pairs of images from the Middlebury \cite{38} dataset for training. We use four benchmark test sets: KITTI 2012 \cite{39}, KITTI 2015 \cite{40}, Middlebury \cite{38}, and Flickr 1024 \cite{37} 
, to fully verify the effectiveness of our model.

\textbf{Evaluation metrics.} We evaluate our model using Peak Signal-To-Noise Ratio (PSNR) and Structural Similarity Index (SSIM) on the RGB color space.

\textbf{Model Setting.} The number of MFF Block and feature channels is flexible and can be changed. In this paper, we adjust the settings in NTIRE 2024 competition to further reduce the number of parameters and improve efficiency. In the ×2 SR model, the number of blocks and the number of channels are set to 16 and 64, respectively. In the ×4 SR model, the number of blocks and the number of channels are set to 24 and 48, respectively.

\textbf{Training Setting.} We augment the training data using random horizontal flipping, rotation, and RGB channel shuffling. We use Lion \cite{41} optimizer with   $\beta_1 = 0.9$, $\beta_2 = 0.99$. MFFSSR is training in PyTorch on a server with eight Nvidia A100 GPUs. The learning rate is initially set to $5 \times {10^{ - 4}}$ and decay the learning rate with the cosine strategy. The total iterations for the model is set to 200,000.

\subsection{Comparisons with State-of-the-art Methods}
In this section, we compare our proposed MFFSSR with existing image SR methods. These methods include VDSR \cite{15}, EDSR \cite{18}, RDN\cite{7}, RCAN\cite{19}, StereoSR \cite{8}, PASSRnet\cite{9}, IMSSRnet \cite{42}, iPASSSR \cite{32}, SSRDE-FNet \cite{36}, PFT-SSR \cite{43}, SwinFIR \cite{5}, NAFSSR \cite{10} and CVHSSR \cite{13} and so on. All models are trained on the same datasets, and these results are from \cite{13}.
\begin{table}
\caption{Ablation experiments of different elements in hybrid attention feature
extraction block on the Flickr1024 test set \cite{37}. PSNR and SSIM are used to evaluate performance.}
\label{tab:3}
\centering
\resizebox{7cm}{!}{%
\begin{tabular}{c|cccccc}
\bottomrule[1.2pt]
         & MFFSSR      & Net-A     & Net-B      & Net-C  & Net-D             \\ \hline
LKA      & \ding{52}    & \ding{56}   & \ding{52}    & \ding{56}     & \ding{52}  \\
RepConv     & \ding{52}    & \ding{52}   & \ding{56}    & \ding{56}     & \ding{52}  \\
IRF      & \ding{52}    & \ding{52}    & \ding{52}   & \ding{52}   & \ding{56}\\
FFN& \ding{56}& \ding{56}& \ding{56}& \ding{56}& \ding{52}\\\hline
PSNR     & 23.92  & 23.87 & 23.91  & 23.85  & 23.91   \\
SSIM     & 0.7503 & 0.7490 & 0.7498 & 0.7476  & 0.7501  \\\hline
$\Delta$ PSNR    & 0      & -0.05  & -0.01    & -0.07 &-0.01    \\ \toprule[1.2pt]
\end{tabular}%
}
\end{table}

\textbf{Quantitative Evaluations.} We summarize the SR results of MFFSSR and other SR methods at both ×2 and ×4 upsampling factors in Table \ref{tab:my-table}. Our MFFSSR achieves the best performance with low parameters.

\textbf{Efficiency Evaluations.} We compare the parameters and computational complexity with the state-of-the-art methods from the NTIRE Stereo Image SR Challenge in 2022 and 2023. We achieve better performance than NAFSSR-S \cite{10} while utilizing only 60\% of the parameters and 75\% of the FLOPs. For fair comparison, we reduce the parameter of SwinFIRSSR \cite{5} and SCGLANet \cite{scglanet} to a level equivalent to that of MFFSSR, resulting in two variant networks: SwinFIRSSR-\textit{v} and SCGLANet-\textit{v}. As shown in Table \ref{tab:2}, even in the condition that the parameters are slightly greater than SwinFIRSSR-\textit{v} and SCGLANet-\textit{v}, our computational cost remains lower. The evaluations are tested on 128×128 size as inputs.

\textbf{Visual Comparison.} Figure \ref{cpimageflickr}. and Figure \ref{cpimagekitti}. display the ×4 SR visualization results of different methods, our model is more visually realistic in perception and provides a clearer restoration of textures and features compared with existing methods. It notably achieves a better restoration effect on fences and buildings.

\begin{table}
\caption{Ablation experiments of different weights $\theta$ and cross-attention modules on the Flickr1024 test set \cite{37}. Params represents the number of parameters of the network. PSNR, SSIM, Params and FLOPs are used to evaluate performance and efficiency.}
\label{tab:4}
\centering
\resizebox{8cm}{!}{%
\begin{tabular}{c|cccccc}
\bottomrule[1.2pt]

         & $\theta  = 0.250$      & $\theta  = 0.500$     & $\theta  = 0.750$          &$\theta  = 0.750$    & $\theta  = 0.875$     \\ \hline
CVIM     & \ding{52}              & \ding{52}             & \ding{52}                                & \ding{56}   & \ding{52}\\
SCAM     & \ding{56}               & \ding{56}            & \ding{56}                         & \ding{52}   & \ding{56}  \\ \hline

PSNR     & 23.89  & 23.90 & 23.92  & 23.89 & 23.86   \\
SSIM     & 0.7489 & 0.7493 & 0.7503 & 0.7487 & 0.7465   \\
Params    & 1.52M      & 1.13M  & 0.84M  & 0.80M  & 0.77M    \\ 
FLOPs     &119.157G   &100.191G   &89.187G  &88.536G &86.671G 
   \\\hline
$\Delta$ PSNR    & -0.03    & -0.02  & 0  & -0.03 & -0.06
\\ 
$\Delta$ Params    & +0.68M    & +0.29M  & 0  & -0.04M & -0.07M
\\
$\Delta$ FLOPs    & +29.970G    & +11.004G  & 0  & -0.651G & -2.516G  
\\ \toprule[1.2pt]
\end{tabular}%
}
\end{table}
\subsection{Ablation Study}
\label{sec:4.3}
In this section, we conduct various ablation experiments to validate the effectiveness of the proposed structures and parameter settings. All ablation results are obtained using the Flickr1024 \cite{37} test set.

\textbf{Effectiveness of Hybrid Attention Feature
Extraction Block.} To further validate the effectiveness of the proposed structures, we investigate the roles of different elements in HAFEB, resulting in four network variants: Net-A (without LKA), Net-B (without RepConv), Net-C (without LKA and RepConv) and Net-D (replace IRF with the simple FFN in NAFSSR \cite{10}). The comparison results are presented in Table \ref{tab:3}. LKA plays the most important role in feature extraction process, which offers a wider receptive field.
RepConv enhances the network's generalization capability and perceptual capacity for details. The HAFEB leverages a residual structure to fully preserve the high and low-level features obtained from different elements, and integrates them with the cross-view features obtained by CVIM. We can observe that the performance of MFFSSR deteriorates when LKA and RepConv are removed. Removing both results in a decrease in PSNR of 0.07 dB. In addition, IRF more effectively regulate the information flow, achieving the 0.01 dB improvement in PSNR compared to the original FFN. These results demonstrate the crucial role these elements play in our network, indicating their indispensability.

\textbf{Performance and Efficiency Trade-offs.}
As shown in Table \ref{tab:4}, we compare the effects of (a) different branch weights $\theta$ in the multi-level extraction and fusion structure and (b) different cross-attention modules on network performance and efficiency. 
As the intra-view feature extraction branch weights $\theta$ gradually increase, the  parameters and FLOPs decrease. However, $\theta$ too big leads to insufficient utilization of complementing information from cross-views, causing bad performance. When $\theta=0.750$, we achieve the best performance, but the increase in parameters and FLOPs is minimal compared to $\theta=0.875$.
Additionally, we compare CVIM with NAFSSR's
cross-attention module, SCAM \cite{10}.  We obtain a 0.03 dB improvement in PSNR at the cost of 0.04M parameters and 0.651G FLOPs. We believe this trade-off is worthwhile.

\subsection{NTIRE Stereo Image SR Challenge}
We have submitted the results of our original model to the NTIRE 2024 Stereo Image Super-Resolution Challenge \cite{NTIRE2024}. Our final scores are 23.53 dB PSNR and 21.50 dB PSNR in Track 1 Constrained SR \& Bicubic Degradation and Track 2 Constrained SR \& Realistic Degradation, respectively, ranking 7th and 9th.

%% file: sec/5_Conclusion.tex
\section{Conclusion}
\label{sec:5}

In this work, we propose the Multi-Level Feature Fusion Network for Lightweight Stereo Image Super-Resolution (MFFSSR). By improving the process of intra-view feature extraction for stereo images, we introduce the hybrid attention feature extraction block, which can effectively extract multi-level features. Furthermore, we creatively integrate the cross-view interaction module into the intra-view feature extraction structure, which greatly improves  effectiveness and decreases the parameters. Extensive experiments demonstrate our proposed model outperforms state-of-the-art methods in stereo image super-resolution.